\PassOptionsToPackage{unicode}{hyperref}
\PassOptionsToPackage{hyphens}{url}
\PassOptionsToPackage{dvipsnames,svgnames,x11names}{xcolor}
\documentclass[
]{article}

\usepackage{amsmath,amssymb}
\usepackage{iftex}
\ifPDFTeX
  \usepackage[T1]{fontenc}
  \usepackage[utf8]{inputenc}
  \usepackage{textcomp} 
\else 
  \usepackage{unicode-math}
  \defaultfontfeatures{Scale=MatchLowercase}
  \defaultfontfeatures[\rmfamily]{Ligatures=TeX,Scale=1}
\fi
\usepackage{lmodern}
\ifPDFTeX\else  
\fi
\IfFileExists{upquote.sty}{\usepackage{upquote}}{}
\IfFileExists{microtype.sty}{
  \usepackage[]{microtype}
  \UseMicrotypeSet[protrusion]{basicmath} 
}{}
\makeatletter
\@ifundefined{KOMAClassName}{
  \IfFileExists{parskip.sty}{%
    \usepackage{parskip}
  }{
    \setlength{\parindent}{0pt}
    \setlength{\parskip}{6pt plus 2pt minus 1pt}}
}{
  \KOMAoptions{parskip=half}}
\makeatother
\usepackage{xcolor}
\ifx\paragraph\undefined\else
  \let\oldparagraph\paragraph
  \renewcommand{\paragraph}[1]{\oldparagraph{#1}\mbox{}}
\fi
\ifx\subparagraph\undefined\else
  \let\oldsubparagraph\subparagraph
  \renewcommand{\subparagraph}[1]{\oldsubparagraph{#1}\mbox{}}
\fi

\providecommand{\tightlist}{%
  \setlength{\itemsep}{0pt}\setlength{\parskip}{0pt}}\usepackage{longtable,booktabs,array}
\usepackage{calc} 
\usepackage{etoolbox}
\makeatletter
\patchcmd\longtable{\par}{\if@noskipsec\mbox{}\fi\par}{}{}
\makeatother
\IfFileExists{footnotehyper.sty}{\usepackage{footnotehyper}}{\usepackage{footnote}}
\makesavenoteenv{longtable}
\usepackage{graphicx}
\makeatletter
\def\maxwidth{\ifdim\Gin@nat@width>\linewidth\linewidth\else\Gin@nat@width\fi}
\def\maxheight{\ifdim\Gin@nat@height>\textheight\textheight\else\Gin@nat@height\fi}
\makeatother
\setkeys{Gin}{width=\maxwidth,height=\maxheight,keepaspectratio}
\makeatletter
\def\fps@figure{htbp}
\makeatother
\newlength{\cslhangindent}
\newlength{\csllabelwidth}
\newlength{\cslentryspacingunit} 
\newenvironment{CSLReferences}[2] 
 {
  \setlength{\parindent}{0pt}
  \ifodd #1
  \let\oldpar\par
  \def\par{\hangindent=\cslhangindent\oldpar}
  \fi
  \setlength{\parskip}{#2\cslentryspacingunit}
 }%
 {}
\usepackage{calc}

\usepackage{listings}
\usepackage{arxiv}
\usepackage{orcidlink}
\usepackage{amsmath}
\usepackage[T1]{fontenc}
\makeatletter
\@ifpackageloaded{caption}{}{\usepackage{caption}}
\AtBeginDocument{%
\ifdefined\contentsname
  \renewcommand*\contentsname{Table of contents}
\else
  \newcommand\contentsname{Table of contents}
\fi
\ifdefined\listfigurename
  \renewcommand*\listfigurename{List of Figures}
\else
  \newcommand\listfigurename{List of Figures}
\fi
\ifdefined\listtablename
  \renewcommand*\listtablename{List of Tables}
\else
  \newcommand\listtablename{List of Tables}
\fi
\ifdefined\figurename
  \renewcommand*\figurename{Figure}
\else
  \newcommand\figurename{Figure}
\fi
\ifdefined\tablename
  \renewcommand*\tablename{Table}
\else
  \newcommand\tablename{Table}
\fi
}
\@ifpackageloaded{float}{}{\usepackage{float}}
\floatstyle{ruled}
\@ifundefined{c@chapter}{\newfloat{codelisting}{h}{lop}}{\newfloat{codelisting}{h}{lop}[chapter]}
\floatname{codelisting}{Listing}

\makeatother
\makeatletter
\@ifpackageloaded{caption}{}{\usepackage{caption}}
\@ifpackageloaded{subcaption}{}{\usepackage{subcaption}}
\makeatother
\makeatletter
\@ifpackageloaded{tcolorbox}{}{\usepackage[skins,breakable]{tcolorbox}}
\makeatother
\makeatletter
\@ifundefined{shadecolor}{\definecolor{shadecolor}{rgb}{.97, .97, .97}}
\makeatother
\makeatletter
\ifdefined\Shaded\renewenvironment{Shaded}{\begin{tcolorbox}[breakable, enhanced, interior hidden, boxrule=0pt, frame hidden, sharp corners, borderline west={3pt}{0pt}{shadecolor}]}{\end{tcolorbox}}\fi
\makeatother
\ifLuaTeX
  \usepackage{selnolig}  
\fi
\IfFileExists{bookmark.sty}{\usepackage{bookmark}}{\usepackage{hyperref}}
\IfFileExists{xurl.sty}{\usepackage{xurl}}{} 
\hypersetup{
  pdftitle={Explaining Black-Box Models through Counterfactuals},
  pdfkeywords={Julia, Explainable Artificial
Intelligence, Counterfactual Explanations, Algorithmic Recourse},
  colorlinks=true,
  linkcolor={blue},
  filecolor={Maroon},
  citecolor={Blue},
  urlcolor={Blue},
  pdfcreator={LaTeX via pandoc}}

\renewcommand{\today}{2023-08-14}
\newcommand{\runninghead}{A Preprint }
\title{Explaining Black-Box Models through Counterfactuals}
\author{
\textbf{Patrick Altmeyer}~\orcidlink{0000-0003-4726-8613}\\\\Delft
University of Technology\\\\\\\\\\
\textbf{Arie van Deursen}\\\\Delft University of Technology\\\\\\\\\\
\textbf{Cynthia C. S. Liem}\\\\Delft University of Technology\\\\}
\date{2023-08-14}
\begin{document}
\maketitle
\begin{abstract}
We present
\href{https://github.com/JuliaTrustworthyAI/CounterfactualExplanations.jl}{\texttt{CounterfactualExplanations.jl}}:
a package for generating Counterfactual Explanations (CE) and
Algorithmic Recourse (AR) for black-box models in Julia. CE explain how
inputs into a model need to change to yield specific model predictions.
Explanations that involve realistic and actionable changes can be used
to provide AR: a set of proposed actions for individuals to change an
undesirable outcome for the better. In this article, we discuss the
usefulness of CE for Explainable Artificial Intelligence and demonstrate
the functionality of our package. The package is straightforward to use
and designed with a focus on customization and extensibility. We
envision it to one day be the go-to place for explaining arbitrary
predictive models in Julia through a diverse suite of counterfactual
generators.
\end{abstract}
{\bfseries \emph Keywords}
\def\sep{\textbullet\ }
Julia \sep Explainable Artificial Intelligence \sep Counterfactual
Explanations \sep 
Algorithmic Recourse

\hypertarget{sec-intro}{%
\section{Introduction}\label{sec-intro}}

Machine Learning models like Deep Neural Networks have become so complex
and opaque over recent years that they are generally considered
black-box systems. This lack of transparency exacerbates several other
problems typically associated with these models: they tend to be
unstable (Goodfellow, Shlens, and Szegedy 2014), encode existing biases
(Buolamwini and Gebru 2018) and learn representations that are
surprising or even counter-intuitive from a human perspective (Sturm
2014). Nonetheless, they often form the basis for data-driven
decision-making systems in real-world applications.

As others have pointed out, this scenario gives rise to an undesirable
principal-agent problem involving a group of principals---i.e.~human
stakeholders---that fail to understand the behaviour of their
agent---i.e.~the black-box system (Borch 2022). The group of principals
may include programmers, product managers and other decision-makers who
develop and operate the system as well as those individuals ultimately
subject to the decisions made by the system. In practice, decisions made
by black-box systems are typically left unchallenged since the group of
principals cannot scrutinize them:

\begin{quote}
``You cannot appeal to (algorithms). They do not listen. Nor do they
bend.'' (O'Neil 2016)
\end{quote}

In light of all this, a quickly growing body of literature on
Explainable Artificial Intelligence (XAI) has emerged. Counterfactual
Explanations fall into this broad category. They can help human
stakeholders make sense of the systems they develop, use or endure: they
explain how inputs into a system need to change for it to produce
different decisions. Explainability benefits internal as well as
external quality assurance. Explanations that involve plausible and
actionable changes can be used for Algorithmic Recourse (AR): they offer
the group of principals a way to not only understand their agent's
behaviour but also adjust or react to it.

The availability of open-source software to explain black-box models
through counterfactuals is still limited. Through the work presented
here, we aim to close that gap and thereby contribute to broader
community efforts towards XAI. We envision this package to one day be
the go-to place for Counterfactual Explanations in Julia. Thanks to
Julia's unique support for interoperability with foreign programming
languages we believe that this library may also benefit the broader
machine learning and data science community.

Our package provides a simple and intuitive interface to generate CE for
many standard classification models trained in Julia, as well as in
Python and R. It comes with detailed documentation involving various
illustrative example datasets, models and counterfactual generators for
binary and multi-class prediction tasks. A carefully designed package
architecture allows for a seamless extension of the package
functionality through custom generators and models.

The remainder of this article is structured as follows:
Section~\ref{sec-related} presents related work on XAI as well as a
brief overview of the methodological framework underlying CE.
Section~\ref{sec-arch} introduces the Julia package and its high-level
architecture. Section~\ref{sec-use} presents several basic and advanced
usage examples. In Section~\ref{sec-custom} we demonstrate how the
package functionality can be customized and extended. To illustrate its
practical usability, we explore examples involving real-world data in
Section~\ref{sec-emp}. Finally, we also discuss the current limitations
of our package, as well as its future outlook in
Section~\ref{sec-outlook}. Section~\ref{sec-conclude} concludes.

\hypertarget{sec-related}{%
\section{Background and related work}\label{sec-related}}

In this section, we first briefly introduce the broad field of
Explainable AI, before narrowing it down to Counterfactual Explanations.
We introduce the methodological framework and finally point to existing
open-source software.

\hypertarget{literature-on-explainable-ai}{%
\subsection{Literature on Explainable
AI}\label{literature-on-explainable-ai}}

The field of XAI is still relatively young and made up of a variety of
subdomains, definitions, concepts and taxonomies. Covering all of these
is beyond the scope of this article, so we will focus only on high-level
concepts. The following literature surveys provide more detail: Arrieta
et al.~(2020) provide a broad overview of XAI (Arrieta et al. 2020); Fan
et al.~(2020) focus on explainability in the context of deep learning
(Fan, Xiong, and Wang 2020); and finally, Karimi et al.~(2020) (Karimi,
Barthe, et al. 2020) and Verma et al.~(2020) Verma, Dickerson, and Hines
(2020) offer detailed reviews of the literature on Counterfactual
Explanations and Algorithmic Recourse (see also Molnar (2020) and
Varshney (2022)). Miller (2019) explicitly discusses the concept of
explainability from the perspective of a social scientist (Miller 2019).

The first broad distinction we want to make here is between
\textbf{Interpretable} and \textbf{Explainable} AI. These terms are
often used interchangeably, but this can lead to confusion. We find the
distinction made in (Rudin 2019) useful: Interpretable AI involves
models that are inherently interpretable and transparent such as general
additive models (GAM), decision trees and rule-based models; Explainable
AI involves models that are not inherently interpretable but require
additional tools to be explainable to humans. Examples of the latter
include Ensembles, Support Vector Machines and Deep Neural Networks.
Some would argue that we best avoid the second category of models
altogether and instead focus solely on interpretable AI (Rudin 2019).
While we agree that initial efforts should always be geared towards
interpretable models, avoiding black boxes altogether would entail
missed opportunities and anyway is probably not very realistic at this
point. For that reason, we expect the need for XAI to persist in the
medium term. Explainable AI can further be broadly divided into
\textbf{global} and \textbf{local} explainability: the former is
concerned with explaining the average behaviour of a model, while the
latter involves explanations for individual predictions (Molnar 2020).
Tools for global explainability include partial dependence plots (PDP),
which involve the computation of marginal effects through Monte Carlo,
and global surrogates. A surrogate model is an interpretable model that
is trained to explain the predictions of a black-box model.

Counterfactual Explanations fall into the category of local methods:
they explain how individual predictions change in response to individual
feature perturbations. Among the most popular alternatives to
Counterfactual Explanations are local surrogate explainers including
Local Interpretable Model-agnostic Explanations (LIME) and Shapley
additive explanations (SHAP). Since explanations produced by LIME and
SHAP typically involve simple feature importance plots, they arguably
rely on reasonably interpretable features at the very least. Contrary to
Counterfactual Explanations, for example, it is not obvious how to apply
LIME and SHAP to high-dimensional image data. Nonetheless, local
surrogate explainers are among the most widely used XAI tools today,
potentially because they are easy to interpret and implemented in
popular programming languages. Proponents of surrogate explainers also
commonly mention that there is a straightforward way to assess their
reliability: a surrogate model that generates predictions in line with
those produced by the black-box model is said to have high
\textbf{fidelity} and therefore considered reliable. As intuitive as
this notion may be, it also points to an obvious shortfall of surrogate
explainers: even a high-fidelity surrogate model that produces the same
predictions as the black-box model 99 per cent of the time is useless
and potentially misleading for every 1 out of 100 individual
predictions.

A recent study has shown that even experienced data scientists tend to
put too much trust in explanations produced by LIME and SHAP (Kaur et
al. 2020). Another recent work has shown that both methods can be easily
fooled: they depend on random input perturbations, a property that can
be abused by adverse agents to essentially whitewash strongly biased
black-box models (Slack et al. 2020). In related work, the same authors
find that while gradient-based Counterfactual Explanations can also be
manipulated, there is a straightforward way to protect against this in
practice (Slack et al. 2021). In the context of quality assessment, it
is also worth noting that---contrary to surrogate explainers---CE always
achieve full fidelity by construction: counterfactuals are searched with
respect to the black-box classifier, not some proxy for it. That being
said, CE should also be used with care and research around them is still
in its early stages.

\hypertarget{sec-method}{%
\subsection{A framework for Counterfactual
Explanations}\label{sec-method}}

Counterfactual search involves feature perturbations: we are interested
in understanding how we need to change individual attributes in order to
change the model output to a desired value or label (Molnar 2020).
Typically the underlying methodology is presented in the context of
binary classification: \(M: \mathcal{X} \mapsto \mathcal{Y}\) where
\(\mathcal{X}\subset\mathbb{R}^D\) and \(\mathcal{Y}=\{0,1\}\). Further,
let \(t=1\) be the target class and let \(x\) denote the factual feature
vector of some individual sample outside of the target class, so
\(y=M(x)=0\). We follow this convention here, though it should be noted
that the ideas presented here also carry over to multi-class problems
and regression (Molnar 2020).

The counterfactual search objective originally proposed by Wachter et
al.~(2017) (Wachter, Mittelstadt, and Russell 2017) is as follows

\begin{equation}\protect\hypertarget{eq-obj}{}{
\min_{x^\prime \in \mathcal{X}} h(x^\prime) \ \ \ \mbox{s. t.} \ \ \ M(x^\prime) = t
}\label{eq-obj}\end{equation}

where \(h(\cdot)\) quantifies how complex or costly it is to go from the
factual \(x\) to the counterfactual \(x^\prime\). To simplify things we
can restate this constrained objective as the following unconstrained
and differentiable problem:

\begin{equation}\protect\hypertarget{eq-solution}{}{
x^\prime = \arg \min_{x^\prime}  \ell(M(x^\prime),t) + \lambda h(x^\prime)
}\label{eq-solution}\end{equation}

Here \(\ell\) denotes some loss function targeting the deviation between
the target label and the predicted label and \(\lambda\) governs the
strength of the complexity penalty. Provided we have gradient access for
the black-box model \(M\) the solution to this problem can be found
through gradient descent. This generic framework lays the foundation for
most state-of-the-art approaches to counterfactual search and is also
used as the baseline approach in our package. The hyperparameter
\(\lambda\) is typically tuned through grid search or in some sense
pre-determined by the nature of the problem. Conventional choices for
\(\ell\) include margin-based losses like cross-entropy loss and hinge
loss. It is worth pointing out that the loss function is typically
computed with respect to logits rather than predicted probabilities, a
convention that we have chosen to follow.\footnote{Implementations of
  loss functions with respect to logits are often numerically more
  stable. For example, the \texttt{logitbinarycrossentropy(ŷ,\ y)}
  implementation in \texttt{Flux.Losses} (used here) is more stable than
  the mathematically equivalent \texttt{binarycrossentropy(ŷ,\ y)}.}

Numerous extensions to this simple approach have been developed since CE
were first proposed in 2017 (see (Verma, Dickerson, and Hines 2020) and
(Karimi, Barthe, et al. 2020) for surveys). The various approaches
largely differ in that they use different flavours of search objective
defined in Equation~\ref{eq-solution}. Different penalties are often
used to address many of the desirable properties of effective CE that
have been set out. These desiderata include: \textbf{proximity} --- the
distance between factual and counterfactual features should be small
(Wachter, Mittelstadt, and Russell 2017); \textbf{actionability} --- the
proposed recourse should be actionable Poyiadzi et al. (2020);
\textbf{plausibility} --- the counterfactual explanation should be
plausible to a human Schut et al. (2021); \textbf{sparsity} --- the
counterfactual explanation should involve as few individual feature
changes as possible (Schut et al. 2021); \textbf{robustness} --- the
counterfactual explanation should be robust to domain and model shifts
(Upadhyay, Joshi, and Lakkaraju 2021); \textbf{diversity} --- ideally
multiple diverse counterfactuals should be provided (Mothilal, Sharma,
and Tan 2020); and \textbf{causality} --- counterfactuals should respect
the structural causal model underlying the data generating process
Karimi, Schölkopf, and Valera (2021).

Beyond gradient-based counterfactual search, which has been the main
focus in our development so far, various methodologies have been
proposed that can handle non-differentiable models like decision trees.
We have implemented some of these approaches and will discuss them
further in Section~\ref{sec-gen}.

\hypertarget{existing-software}{%
\subsection{Existing software}\label{existing-software}}

To the best of our knowledge, the package introduced here provides the
first implementation of Counterfactual Explanations in Julia and
therefore represents a novel contribution to the community. As for other
programming languages, we are only aware of one other unifying
framework: the Python library
\href{https://carla-counterfactual-and-recourse-library.readthedocs.io/en/latest/?badge=latest}{CARLA}
(Pawelczyk et al. 2021).\footnote{While we were writing this paper, the
  \texttt{R} package \texttt{counterfactuals} was released (Dandl et al.
  2023). The developers seem to also envision a unifying framework, but
  the project appears to still be in its early stages.} In addition to
that, there exists open-source code for some specific approaches to CE
that have been proposed in recent years. The approach-specific
implementations that we have been able to find are generally
well-documented, but exclusively in Python. For example, a PyTorch
implementation of a greedy generator for Bayesian models proposed in
(Schut et al. 2021) has been released. As another example, the popular
\href{https://github.com/interpretml}{InterpretML} library includes an
implementation of a diverse counterfactual generator (Mothilal, Sharma,
and Tan 2020).

Generally speaking, software development in the space of XAI has largely
focused on various global methods and surrogate explainers:
implementations of PDP, LIME and SHAP are available for both Python
(e.g.~\href{https://github.com/marcotcr/lime}{\texttt{lime}},
\href{https://github.com/slundberg/shap}{\texttt{shap}}) and R
(e.g.~\href{https://cran.r-project.org/web/packages/lime/index.html}{\texttt{lime}},
\href{https://cran.r-project.org/web/packages/lime/index.html}{\texttt{iml}},
\href{https://modeloriented.github.io/shapper/}{\texttt{shapper}},
\href{https://github.com/bgreenwell/fastshap}{\texttt{fastshap}}). In
the Julia space, there exist two packages related to XAI: firstly,
\href{https://github.com/nredell/ShapML.jl}{\texttt{ShapML.jl}}, which
provides a fast implementation of SHAP; and, secondly,
\href{https://github.com/adrhill/ExplainableAI.jl}{\texttt{ExplainableAI.jl}},
which enables users to easily visualise gradients and activation maps
for \texttt{Flux.jl} models. We also should not fail to mention the
comprehensive
\href{https://docs.interpretable.ai/stable/IAIBase/data/}{Interpretable
AI} infrastructure, which focuses exclusively on interpretable models.

Arguably the current availability of tools for explaining black-box
models in Julia is limited, but it appears that the community is
invested in changing that. The team behind \texttt{MLJ.jl}, for example,
recruited contributors for a project about both Interpretable and
Explainable AI in 2022.\footnote{For details, see the Google Summer of
  Code 2022 project proposal:
  \url{https://julialang.org/jsoc/gsoc/MLJ/\#interpretable_machine_learning_in_julia}.}
With our work on Counterfactual Explanations we hope to contribute to
these efforts. We think that because of its unique transparency the
Julia language naturally lends itself towards building Trustworthy AI
systems.

\hypertarget{sec-arch}{%
\section{\texorpdfstring{Introducing:
\texttt{CounterfactualExplanations.jl}}{Introducing: CounterfactualExplanations.jl}}\label{sec-arch}}

\begin{figure}

{\centering \includegraphics[width=3.33333in,height=2.38095in]{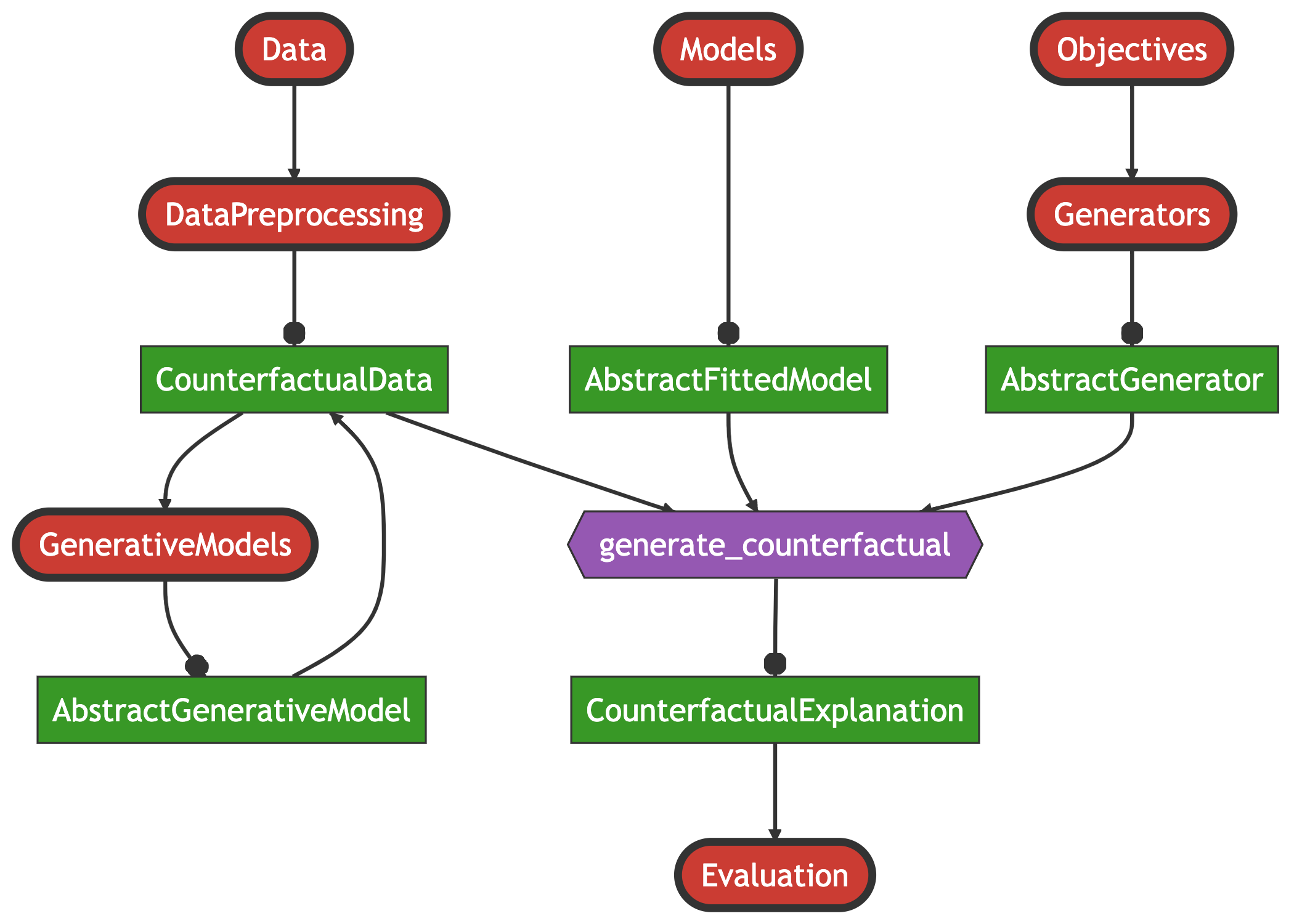}

}

\caption{\label{fig-arch}High-level schematic overview of package
architecture. Modules are shown in red, structs in green and functions
in purple.}

\end{figure}

Figure~\ref{fig-arch} provides an overview of the package architecture.
It is built around two core modules that are designed to be as
extensible as possible through dispatch: 1) \texttt{Models} is concerned
with making any arbitrary model compatible with the package; 2)
\texttt{Generators} is used to implement counterfactual search
algorithms. The core function of the
package---\texttt{generate\_counterfactual}---uses an instance of type
\texttt{\textless{}:AbstractFittedModel} produced by the \texttt{Models}
module and an instance of type \texttt{\textless{}:AbstractGenerator}
produced by the \texttt{Generators} module. Relating this to the
methodology outlined in Section~\ref{sec-method}, the former instance
corresponds to the model \(M\), while the latter defines the rules for
the counterfactual search (Equation~\ref{eq-solution}).

\hypertarget{models}{%
\subsection{Models}\label{models}}

The package currently offers native support for models built and trained
in \href{https://fluxml.ai/}{Flux} (Innes 2018) as well as a small
subset of models made available through
\href{https://alan-turing-institute.github.io/MLJ.jl/dev/}{MLJ} (Blaom
et al. 2020). While in general it is assumed that users resort to this
package to explain their pre-trained models, we provide a simple API
call to train the following
\href{https://juliatrustworthyai.github.io/CounterfactualExplanations.jl/v0.1/tutorials/model_catalogue/}{models}:

\begin{itemize}
\tightlist
\item
  Linear Classifier (Logistic Regression and Multinomial Logit)
\item
  Multi-Layer Perceptron (Deep Neural Network)
\item
  Deep Ensemble (Lakshminarayanan, Pritzel, and Blundell 2016)
\item
  Decision Tree, Random Forest, Gradient Boosted Trees
\end{itemize}

As we demonstrate below, it is straightforward to extend the package
through custom models. Support for \texttt{torch} models trained in
Python or R is also available.\footnote{We are currently relying on
  \texttt{PythonCall.jl} and \texttt{RCall.jl} and this functionality is
  still somewhat brittle. Since this is more of an edge case, we may
  move this feature into its own package in the future.}

\hypertarget{sec-gen}{%
\subsection{Generators}\label{sec-gen}}

A large and growing number of counterfactual
\href{https://juliatrustworthyai.github.io/CounterfactualExplanations.jl/v0.1/explanation/generators/overview/}{generators}
have already been implemented in our package (Table \ref{tab-gen}). At a
high level, we distinguish generators in terms of their compatible model
types, their default search space, and their composability. All
``gradient-based'' generators are compatible with differentiable models,
e.g.~\texttt{Flux} and \texttt{torch}, while ``tree-based'' generators
are only applicable to models that involve decision trees. Concerning
the search space, it is possible to search counterfactuals in a
lower-dimensional latent embedding of the feature space that implicitly
encodes the data-generating process (DGP). To learn the latent
embedding, existing work has typically relied on generative models or
existing causal knowledge Karimi, Schölkopf, and Valera (2021). While
this notion is compatible with all of our gradient-based generators,
only some generators search a latent space by default. Finally,
composability implies that the given generator can be blended with any
other composable generator, which we discuss in
Section~\ref{sec-gen-comp}.

Beyond these broad technical distinctions, generators largely differ in
terms of how they address the various desiderata mentioned above:
\emph{ClapROAR} aims to preserve the classifier, i.e.~to generate
counterfactuals that are robust to endogenous model shifts (Altmeyer et
al. 2023); \emph{CLUE} searches plausible counterfactuals in the latent
embedding of a generative model by explicitly minimising predictive
entropy (Antorán et al. 2020); \emph{DiCE} is designed to generate
multiple, maximally diverse counterfactuals (Mothilal, Sharma, and Tan
2020); \emph{FeatureTweak} leverages the internals of decision trees to
search counterfactuals on a feature-by-feature basis, finding the
counterfactual that tweaks the features in the least costly way (Tolomei
et al. 2017); \emph{Gravitational} aims to generate plausible and robust
counterfactuals by minimising the distance to observed samples in the
target class (Altmeyer et al. 2023); \emph{Greedy} aims to generate
plausible counterfactuals by implicitly minimising predictive
uncertainty of Bayesian classifiers (Schut et al. 2021);
\emph{GrowingSpheres} is model-agnostic, relying solely on identifying
nearest neighbours of counterfactuals in the target class by gradually
increasing the search radius and then moving counterfactuals in that
direction (Laugel et al. 2017); \emph{PROBE} generates probabilistically
robust counterfactuals (Pawelczyk et al. 2022); \emph{REVISE} addresses
the need for plausibility by searching counterfactuals in the latent
embedding of a Variational Autoencoder (VAE) (Joshi et al. 2019);
\emph{Wachter} is the baseline approach that only penalises the distance
to the original sample (Wachter, Mittelstadt, and Russell 2017).

\begin{table}
\caption{Overview of implemented counterfactual generators. \label{tab-gen} \newline}
\centering
\fontsize{7}{9}\selectfont
\begin{tabular}[t]{llll}
\toprule
Generator & Model Type & Search Space & Composable\\
\midrule
ClaPROAR & gradient based & feature & yes\\
CLUE & gradient based & latent & yes\\
DiCE & gradient based & feature & yes\\
FeatureTweak & tree based & feature & no\\
Gravitational & gradient based & feature & yes\\
Greedy & gradient based & feature & yes\\
GrowingSpheres & agnostic & feature & no\\
PROBE & gradient based & feature & no\\
REVISE & gradient based & latent & yes\\
Wachter & gradient based & feature & yes\\
\bottomrule
\end{tabular}
\end{table}

\hypertarget{data-catalogue}{%
\subsection{Data Catalogue}\label{data-catalogue}}

To allow researchers and practitioners to test and compare
counterfactual generators, the package ships with
\href{https://juliatrustworthyai.github.io/CounterfactualExplanations.jl/v0.1/tutorials/data_catalogue/}{catalogues}
of pre-processed synthetic and real-world benchmark datasets from
different domains. Real-world datasets include:

\begin{itemize}
\tightlist
\item
  Adult Census (Barry Becker 1996)
\item
  California Housing (Pace and Barry 1997)
\item
  CIFAR10 (Krizhevsky 2009)
\item
  German Credit (Hoffman 1994)
\item
  Give Me Some Credit (Kaggle 2011)
\item
  MNIST (LeCun 1998) and Fashion MNIST (Xiao, Rasul, and Vollgraf 2017)
\item
  UCI defaultCredit (Yeh and Lien 2009)
\end{itemize}

Custom datasets can also be easily preprocessed as explained in the
\href{https://juliatrustworthyai.github.io/CounterfactualExplanations.jl/v0.1/tutorials/data_preprocessing/}{documentation}.

\hypertarget{plotting}{%
\subsection{Plotting}\label{plotting}}

The package also extends common \texttt{Plots.jl} methods to facilitate
the visualization of results. Calling the generic \texttt{plot()} method
on an instance of type \texttt{\textless{}:CounterfactualExplanation},
for example, generates a plot visualizing the entire counterfactual path
in the feature space\footnote{For multi-dimensional input data, standard
  dimensionality reduction techniques are used to compress the data. In
  this case, the classifier's decision boundary is approximated through
  a Nearest Neighbour model. This is still somewhat experimental and
  will be improved in the future.}. We will see several examples of this
below.

\hypertarget{sec-use}{%
\section{Basic Usage}\label{sec-use}}

In the following, we begin our exploration of the package functionality
with a simple example. We then demonstrate how more advanced generators
can be easily composed and show how users can impose mutability
constraints on features. Finally, we also briefly explore the topics of
counterfactual evaluation and benchmarking.

\hypertarget{sec-simple}{%
\subsection{A Simple Generic Generator}\label{sec-simple}}

\begin{figure}

{\centering \includegraphics[width=3.33333in,height=2.5in]{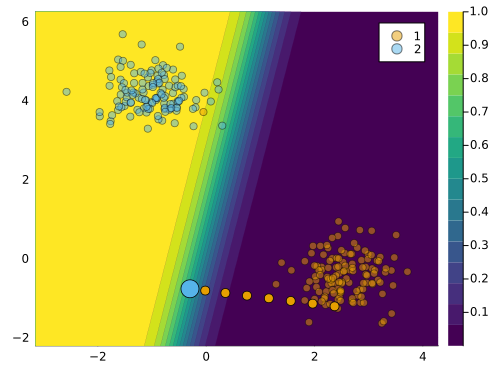}

}

\caption{\label{fig-binary}Counterfactual path using generic
counterfactual generator for conventional binary classifier.}

\end{figure}

Listing \ref{lst:simple} below provides a complete example demonstrating
how the framework presented in Section~\ref{sec-method} can be
implemented through our package. Using a synthetic data set with
linearly separable features we first fit a linear classifier (line
\ref{line:simple-class}). Next, we define the target class (line
\ref{line:simple-t}) and then draw a random sample from the other class
(line \ref{line:simple-x}). Finally, we instantiate a generic generator
(line \ref{line:simple-gen}) and run the counterfactual search (line
\ref{line:simple-search}). Figure~\ref{fig-binary} illustrates the
resulting counterfactual path in the two-dimensional feature space.
Features go through iterative perturbations until the desired confidence
level is reached as illustrated by the contour in the background, which
shows the softmax output for the target class.

\begin{lstlisting}[escapechar=@, numbers=left, label={lst:simple}, caption={Standard workflow for generating counterfactuals.}] 
# Data and Classifier:
counterfactual_data = load_linearly_separable()
M = fit_model(counterfactual_data, :Linear) @\label{line:simple-class}@

# Factual and Target:
yhat = predict_label(M, counterfactual_data)
target = 2    # target label @\label{line:simple-t}@
candidates = findall(vec(yhat) .!= target)
chosen = rand(candidates)
x = select_factual(counterfactual_data, chosen) @\label{line:simple-x}@

# Counterfactual search:
generator = GenericGenerator() @\label{line:simple-gen}@
ce = generate_counterfactual(
    x, target, counterfactual_data, M, generator) @\label{line:simple-search}@
\end{lstlisting}

In this simple example, the generic generator produces a valid
counterfactual, since the decision boundary is crossed and the predicted
label is flipped. But the counterfactual is not plausible: it does not
appear to be generated by the same DGP as the observed data in the
target class. This is because the generic generator does not take into
account any of the desiderata mentioned in Section~\ref{sec-method},
except for the distance to the factual sample.

\hypertarget{sec-gen-comp}{%
\subsection{Composing Generators}\label{sec-gen-comp}}

To address these issues, we can leverage the ideas underlying some of
the more advanced counterfactual generators introduced above. In
particular, we will now show how easy it is to
\href{https://juliatrustworthyai.github.io/CounterfactualExplanations.jl/v0.1/tutorials/generators/}{compose
custom generators} that blend different ideas through user-friendly
macros.

Suppose we wanted to address the desiderata of plausibility and
diversity. We could do this by blending ideas underlying the \emph{DiCE}
generator with the \emph{REVISE} generator. Formally, the corresponding
search objective would be defined as follows,

\begin{equation}\protect\hypertarget{eq-comp}{}{
\mathbf{Z}^\prime = \arg \min_{\mathbf{Z}^\prime \in \mathcal{Z}^{L \times K}} \{  {\ell(M(f(\mathbf{Z}^\prime)),t)} + \lambda \cdot {\text{diversity}(f(\mathbf{Z}^\prime)) }  \} 
}\label{eq-comp}\end{equation}

where \(\mathbf{X}^\prime\) is an \(L\)-dimensional array of
counterfactuals,
\(f: \mathcal{Z}^{L \times K} \mapsto \mathcal{X}^{L \times D}\) is a
function that maps the \(L \times K\)-dimensional latent space
\(\mathcal{Z}\) to the \(L \times D\)-dimensional feature space
\(\mathcal{X}\) and \(\text{diversity}(\cdot)\) is the penalty proposed
by Mothilal et al.~(2020) (Mothilal, Sharma, and Tan 2020) that induces
diverse sets of counterfactuals. As in Equation~\ref{eq-solution},
\(\ell\) is the loss function, \(M\) is the black-box model, \(t\) is
the target class, and \(\lambda\) is the strength of the penalty.

Listing \ref{lst:composed} demonstrates how Equation~\ref{eq-comp} can
be seamlessly translated into Julia code. We begin by instantiating a
\texttt{GradientBasedGenerator} in line \ref{line:composed-init}. Next,
we use chained macros for composition: firstly, we define the
counterfactual search \texttt{@objective} corresponding to \emph{DiCE}
in line \ref{line:composed-dice}; secondly, we define the latent space
search strategy corresponding to \emph{REVISE} using the
\texttt{@search\_latent\_space} macro in line
\ref{line:composed-latent}; finally, we specify our prefered
optimisation method using the \texttt{@with\_optimiser} macro in line
\ref{line:composed-adam}.

\begin{lstlisting}[escapechar=§, numbers=left, label={lst:composed}, caption={Composing a custom generator.}]
generator = GradientBasedGenerator() §\label{line:composed-init}§
@chain generator begin
    @objective logitcrossentropy 
      + 0.2ddp_diversity §\label{line:composed-dice}§
    @search_latent_space §\label{line:composed-latent}§
    @with_optimiser Adam(0.005) §\label{line:composed-adam}§
end
\end{lstlisting}

In this case, the counterfactual search is performed in the latent space
of a Variational Autoencoder (VAE) that is automatically trained on the
observed data. It is important to specify the keyword argument
\texttt{num\_counterfactuals} of the \texttt{generate\_counterfactual}
to some value higher than \(1\) (default), to ensure that the diversity
penalty is effective. The resulting counterfactual path is shown in
Figure~\ref{fig-binary-advanced} below. We observe that the resulting
counterfactuals are diverse and the majority of them are plausible.

\begin{figure}

{\centering \includegraphics[width=3.33333in,height=2.5in]{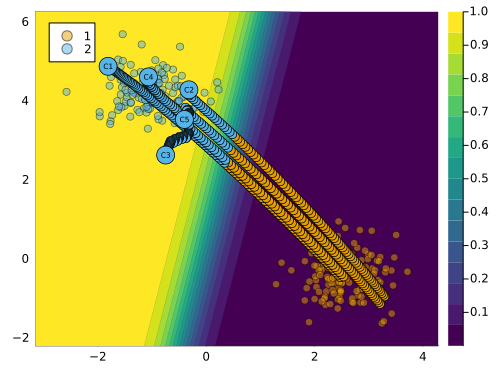}

}

\caption{\label{fig-binary-advanced}Counterfactual path using the
\emph{DiCE} generator.}

\end{figure}

\hypertarget{sec-mut}{%
\subsection{Mutability Constraints}\label{sec-mut}}

In practice, features usually cannot be perturbed arbitrarily. Suppose,
for example, that one of the features used by a bank to predict the
creditworthiness of its clients is \emph{age}. If a counterfactual
explanation for the prediction model indicates that older clients should
``grow younger'' to improve their creditworthiness, then this is an
interesting insight (it reveals age bias), but the provided recourse is
not actionable. In such cases, we may want to constrain the mutability
of features. To illustrate how this can be implemented in our package,
we will continue with the example from above.

Mutability can be defined in terms of four different options: 1) the
feature is mutable in both directions, 2) the feature can only increase
(e.g.~\emph{age}), 3) the feature can only decrease (e.g.~\emph{time
left} until your next deadline) and 4) the feature is not mutable
(e.g.~\emph{skin colour}, \emph{ethnicity}, \ldots). To specify which
category a feature belongs to, users can pass a vector of symbols
containing the mutability constraints at the pre-processing stage. For
each feature one can choose from these four options: \texttt{:both}
(mutable in both directions), \texttt{:increase} (only up),
\texttt{:decrease} (only down) and \texttt{:none} (immutable). By
default, \texttt{nothing} is passed to that keyword argument and it is
assumed that all features are mutable in both directions.\footnote{Mutability
  constraints are not yet implemented for Latent Space search.}

We can impose that the first feature is immutable as follows:
\texttt{counterfactual\_data.mutability\ =\ {[}:none,\ :both{]}}. The
resulting counterfactual path is shown in Figure~\ref{fig-mutability}
below. Since only the second feature can be perturbed, the sample can
only move along the vertical axis. In this case, the counterfactual
search does not yield a valid counterfactual, since the target class is
not reached.

\begin{figure}

{\centering \includegraphics[width=3.33333in,height=2.5in]{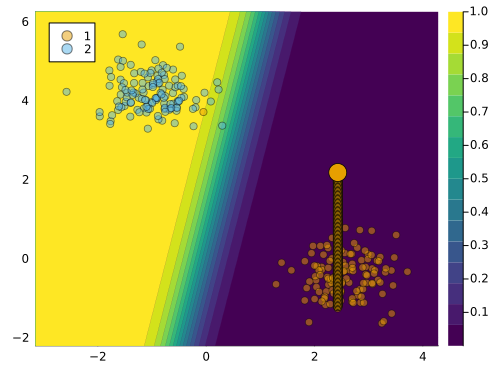}

}

\caption{\label{fig-mutability}Counterfactual path with immutable
feature.}

\end{figure}

\hypertarget{sec-eval}{%
\subsection{Evaluation and Benchmarking}\label{sec-eval}}

The package also makes it easy to
\href{https://juliatrustworthyai.github.io/CounterfactualExplanations.jl/v0.1/tutorials/evaluation/}{evaluate}
counterfactuals with respect to many of the desiderata mentioned above.
For example, users may want to infer how costly the provided recourse is
to individuals. To this end, we can measure the distance of the
counterfactual from its original value. The API call to compute the
distance metric defined in Wachter et al.~(2017) Wachter, Mittelstadt,
and Russell (2017), for instance, is as simple as
\texttt{evaluate(ce;\ measure=distance\_mad)}, where \texttt{ce} can
also be a vector of \texttt{CounterfactualExplanation}s.

\begin{figure}

{\centering \includegraphics[width=3.33333in,height=2.5in]{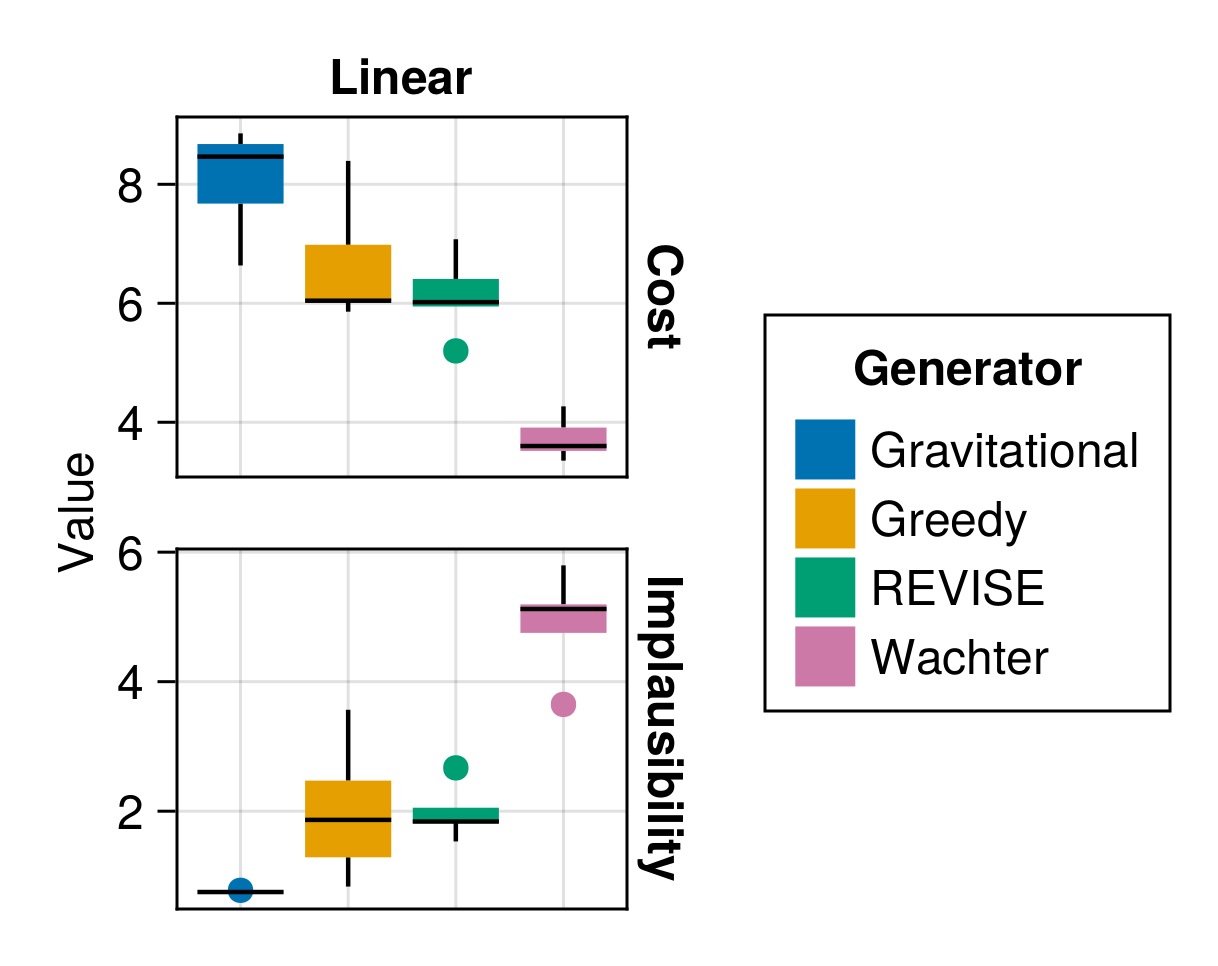}

}

\caption{\label{fig-bmk}Benchmarking results for different generators.}

\end{figure}

Additionally, the package provides a
\href{https://juliatrustworthyai.github.io/CounterfactualExplanations.jl/v0.1/tutorials/benchmarking/}{benchmarking}
framework that allows users to compare the performance of different
generators on a given dataset. In Figure~\ref{fig-bmk} we show the
results of a benchmark comparing several generators in terms of the
average cost and implausibility of the generated counterfactuals. The
cost is proxied by the L1-norm of the difference between the factual and
counterfactual features, while implausibility is measured by the
distance of the counterfactuals from samples in the target class. The
results illustrate that there is a tradeoff between minimizing costs to
individuals and generating plausible counterfactuals.

\hypertarget{sec-custom}{%
\section{Customization and Extensibility}\label{sec-custom}}

One of our priorities has been to make our package customizable and
extensible. In the long term, we aim to add support for more default
models and counterfactual generators. In the short term, it is designed
to allow users to integrate models and generators themselves. These
community efforts will facilitate our long-term goals.

\hypertarget{sec-custom-mod}{%
\subsection{Adding Custom Models}\label{sec-custom-mod}}

At the high level, only two steps are necessary to make any supervised
learning model compatible with our package:

\begin{itemize}
\tightlist
\item
  \textbf{Subtyping}: We need to subtype the
  \texttt{AbstractFittedModel}.
\item
  \textbf{Dispatch}: The functions \texttt{logits} and \texttt{probs}
  need to be extended through custom methods for the model in question.
\end{itemize}

To demonstrate how this can be done in practice, we will reiterate here
how native support for \href{https://fluxml.ai/}{\texttt{Flux.jl}}
(Innes 2018) deep learning models was enabled.\footnote{Flux models are
  now natively supported by our package and can be instantiated by
  calling \texttt{FluxModel()}.} Once again we use synthetic data for an
illustrative example. Listing \ref{lst:nn} below builds a simple model
architecture that can be used for a multi-class prediction task. Note
how outputs from the final layer are not passed through a softmax
activation function, since the counterfactual loss is evaluated with
respect to logits as we discussed earlier. The model is trained with
dropout.

\begin{lstlisting}[escapechar=@, numbers=left, label={lst:nn}, caption={A simple neural network model.}]
n_hidden = 32
output_dim = length(unique(y))
input_dim = 2
model = Chain(
    Dense(input_dim, n_hidden, activation),
    Dropout(0.1),
    Dense(n_hidden, output_dim)
)  
\end{lstlisting}

Listing \ref{lst:mymodel} below implements the two steps that were
necessary to make Flux models compatible with the package. In line
\ref{line:mymodel-subtype} we declare our new struct as a subtype of
\texttt{AbstractDifferentiableModel}, which itself is an abstract
subtype of \texttt{AbstractFittedModel}.\footnote{Note that in line
  \ref{line:mymodel-likelihood} we also provide a field determining the
  likelihood. This is optional and only used internally to determine
  which loss function to use in the counterfactual search. If this field
  is not provided to the model, the loss function needs to be explicitly
  supplied to the generator.} Computing logits amounts to just calling
the model on inputs. Predicted probabilities for labels can be computed
by passing logits through the softmax function.

\begin{lstlisting}[escapechar=@, numbers=left, label={lst:mymodel}, caption={A wrapper for Flux models.}]
# Step 1)
struct MyFluxModel <: AbstractDifferentiableModel @\label{line:mymodel-subtype}@
    model::Any
    likelihood::Symbol @\label{line:mymodel-likelihood}@
end

# Step 2)
# import functions in order to extend
import CounterfactualExplanations.Models: logits
import CounterfactualExplanations.Models: probs 
logits(M::MyFluxModel, X::AbstractArray) = M.model(X)
probs(M::MyFluxModel, X::AbstractArray) = softmax(logits(M, X))
M = MyFluxModel(model)
\end{lstlisting}

\begin{figure}

{\centering \includegraphics[width=3.33333in,height=2.5in]{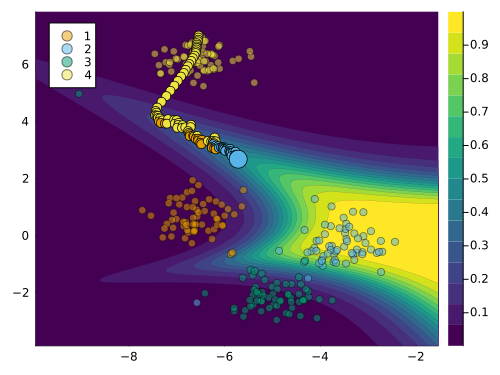}

}

\caption{\label{fig-multi}Counterfactual path using generic
counterfactual generator for multi-class classifier.}

\end{figure}

The API call for generating counterfactuals for our new model is the
same as before. Figure~\ref{fig-multi} shows the resulting
counterfactual path for a randomly chosen sample. In this case, the
contour shows the predicted probability that the input is in the target
class (\(t=2\)).

\hypertarget{sec-custom-gen}{%
\subsection{Adding Custom Generators}\label{sec-custom-gen}}

In some cases, composability may not be sufficient to implement specific
logics underlying certain counterfactual generators. In such cases,
users may want to implement custom generators. To illustrate how this
can be done we will consider a simple extension of our
\texttt{GenericGenerator}. As we have seen above, Counterfactual
Explanations are not unique. In light of this, we might be interested in
quantifying the uncertainty around the generated counterfactuals
(Delaney, Greene, and Keane 2021). One idea could be, to use dropout to
randomly switch features on and off in each iteration. Without dwelling
further on the merit of this idea, we will now briefly show how this can
be implemented.

\hypertarget{a-generator-with-dropout}{%
\subsubsection{A Generator with
Dropout}\label{a-generator-with-dropout}}

Listing \ref{lst:dropout} below implements two important steps: 1)
create an abstract subtype of the
\texttt{AbstractGradientBasedGenerator} and 2) create a constructor with
an additional field for the dropout probability.

\begin{lstlisting}[escapechar=@, numbers=left, label={lst:dropout}, caption={Building a custom generator with dropout.}]
# Abstract suptype:
abstract type AbstractDropoutGenerator <: AbstractGradientBasedGenerator end
# Constructor:
struct DropoutGenerator <: AbstractDropoutGenerator
    loss::Symbol # loss function
    complexity::Function # complexity function
    @$\lambda$@::AbstractFloat # strength of penalty
    decision_threshold::Union{Nothing,AbstractFloat} 
    opt::Any # optimizer
    @$\tau$@::AbstractFloat # tolerance for convergence
    p_dropout::AbstractFloat # dropout rate
end
\end{lstlisting}

Next, in Listing \ref{lst:generate} we define how feature perturbations
are generated for our custom dropout generator: in particular, we extend
the relevant function through a method that implements the dropout
logic.

\begin{lstlisting}[escapechar=@, numbers=left, label={lst:generate}, caption={Generating feature perturbations with dropout.}]
using CounterfactualExplanations.Generators
function Generators.generate_perturbations(
    generator::AbstractDropoutGenerator, 
    ce::CouterfactualExplanation
)
    @$s^\prime$@ = deepcopy(ce.@$s^\prime$@)
    new_@$s^\prime$@ = Generators.propose_state(
        generator, ce)
    @$\Delta s^\prime$@ = new_@$s^\prime$@ - @$s^\prime$@ # gradient step
    # Dropout:
    set_to_zero = sample(
        1:length(@$\Delta s^\prime$@),
        Int(round(generator.p_dropout*length(@$\Delta s^\prime$@))),
        replace=false
    )
    @$\Delta s^\prime$@[set_to_zero] .= 0
    return @$\Delta s^\prime$@
end
\end{lstlisting}

\begin{figure}

{\centering \includegraphics[width=3.33333in,height=2.5in]{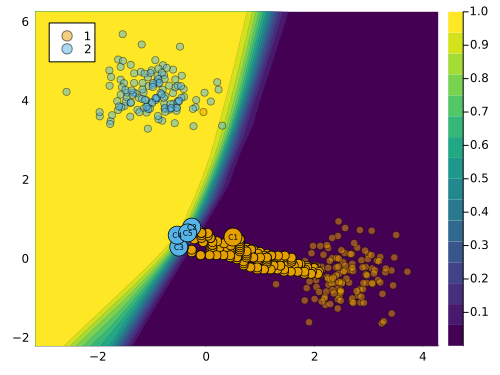}

}

\caption{\label{fig-dropout}Counterfactual path for a generator with
dropout.}

\end{figure}

Finally, we proceed to generate counterfactuals in the same way we
always do. The resulting counterfactual path is shown in
Figure~\ref{fig-dropout}.

\hypertarget{sec-emp}{%
\section{A Real-World Examples}\label{sec-emp}}

Now that we have explained the basic functionality of
\texttt{CounterfactualExplanations.jl} through some synthetic examples,
it is time to work through examples involving real-world data.

\hypertarget{give-me-some-credit}{%
\subsection{Give Me Some Credit}\label{give-me-some-credit}}

The \emph{Give Me Some Credit} dataset is one of the tabular real-world
datasets that ship with the package (Kaggle 2011). It can be used to
train a binary classifier to predict whether a borrower is likely to
experience financial difficulties in the next two years. In particular,
we have an output variable
\(y \in \{0=\texttt{no stress},1=\texttt{stress}\}\) and a feature
matrix \(X\) that includes socio-demographic variables like \texttt{age}
and \texttt{income}. A retail bank might use such a classifier to
determine if potential borrowers should receive credit or not.

For the classification task, we use a Multi-Layer Perceptron with
dropout regularization. Using the Gravitational generator Altmeyer et
al. (2023) we will generate counterfactuals for ten randomly chosen
individuals that would be denied credit based on our pre-trained model.
Concerning the mutability of features, we only impose that the
\texttt{age} cannot be decreased.

\begin{figure}

{\centering \includegraphics[width=3.33333in,height=1.33333in]{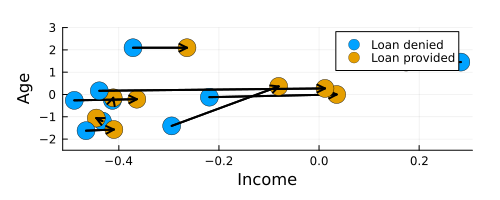}

}

\caption{\label{fig-credit}Give Me Some Credit: counterfactuals for
would-be borrowers proposed by the Gravitational Generator.}

\end{figure}

Figure~\ref{fig-credit} shows the resulting counterfactuals proposed by
Wachter in the two-dimensional feature space spanned by the \texttt{age}
and \texttt{income} variables. An increase in income and age is
recommended for the majority of individuals, which seems plausible: both
age and income are typically positively related to creditworthiness.

\hypertarget{mnist}{%
\subsection{MNIST}\label{mnist}}

\begin{figure}

{\centering \includegraphics[width=3.33333in,height=1.11111in]{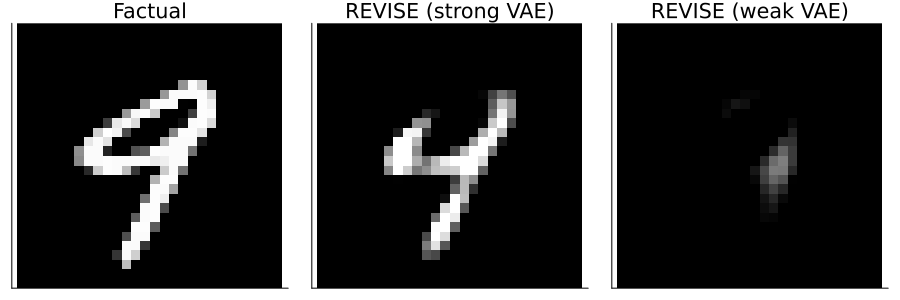}

}

\caption{\label{fig-mnist}Counterfactual explanations for MNIST using a
Latent Space generator: turning a nine (9) into a four (4).}

\end{figure}

For our second example, we will look at image data. The MNIST dataset
contains 60,000 training samples of handwritten digits in the form of
28x28 pixel grey-scale images (LeCun 1998). Each image is associated
with a label indicating the digit (0-9) that the image represents. The
data makes for an interesting case study of CE because humans have a
good idea of what plausible counterfactuals of digits look like. For
example, if you were asked to pick up an eraser and turn the digit in
the left panel of Figure~\ref{fig-mnist} into a four (4) you would know
exactly what to do: just erase the top part.

On the model side, we will use a simple multi-layer perceptron (MLP).
Listing \ref{lst:mnist-setup} loads the data and the pre-trained MLP. It
also loads two pre-trained Variational Auto-Encoders, which will be used
by our counterfactual generator of choice for this task: \emph{REVISE}.

\begin{lstlisting}[escapechar=@, numbers=left, label={lst:mnist-setup}, caption={Loading pre-trained models and data for MNIST.}]
counterfactual_data = load_mnist()
X, y = unpack_data(counterfactual_data)
input_dim, n_obs = size(counterfactual_data.X)
M = load_mnist_mlp()
vae = load_mnist_vae()
vae_weak = load_mnist_vae(;strong=false)
\end{lstlisting}

The proposed counterfactuals are shown in Figure~\ref{fig-mnist}. In the
case in which \emph{REVISE} has access to an expressive VAE (centre),
the result looks convincing: the perturbed image does look like it
represents a four (4). In terms of explainability, we may conclude that
removing the top part of the handwritten nine (9) leads the black-box
model to predict that the perturbed image represents a four (4). We
should note, however, that the quality of counterfactuals produced by
\emph{REVISE} hinges on the performance of the underlying generative
model, as demonstrated by the result on the right. In this case,
\emph{REVISE} uses a weak VAE and the resulting counterfactual is
invalid. In light of this, we recommend using Latent Space search with
care.

\hypertarget{sec-outlook}{%
\section{Discussion and Outlook}\label{sec-outlook}}

We believe that this package in its current form offers a valuable
contribution to ongoing efforts towards XAI in Julia. That being said,
there is significant scope for future developments, which we briefly
outline in this final section.

\hypertarget{candidate-models-and-generators}{%
\subsection{Candidate models and
generators}\label{candidate-models-and-generators}}

The package supports various models and generators either natively or
through minimal augmentation. In future work, we would like to
prioritize the addition of further predictive models and generators.
Concerning the former, it would be useful to add native support for any
supervised models built in \texttt{MLJ.jl}, an extensive Machine
Learning framework for Julia (Blaom et al. 2020). This may also involve
adding support for regression models as well as additional
non-differentiable models. In terms of counterfactual generators, there
is a list of recent methodologies that we would like to implement
including MINT (Karimi, Schölkopf, and Valera 2021), ROAR (Upadhyay,
Joshi, and Lakkaraju 2021) and FACE (Poyiadzi et al. 2020).

\hypertarget{additional-datasets}{%
\subsection{Additional datasets}\label{additional-datasets}}

For benchmarking and testing purposes it will be crucial to add more
datasets to our library. We have so far prioritized tabular datasets
that have typically been used in the literature on counterfactual
explanations including \emph{Adult}, \emph{Give Me Some Credit} and
\emph{German Credit} (Karimi, Barthe, et al. 2020). There is scope for
adding data sources that have so far not been explored much in this
context including additional image datasets as well as audio, natural
language and time-series data.

\hypertarget{sec-conclude}{%
\section{Concluding remarks}\label{sec-conclude}}

\texttt{CounterfactualExplanation.jl} is a package for generating
Counterfactual Explanations and Algorithmic Recourse in Julia. Through
various synthetic and real-world examples, we have demonstrated the
basic usage of the package as well as its extensibility. The package has
already served us in our research to benchmark various methodological
approaches to Counterfactual Explanations and Algorithmic Recourse. We
therefore strongly believe that it should help other practitioners and
researchers in their own efforts towards Trustworthy AI.

We envision this package to one day constitute the go-to place for
explaining arbitrary predictive models through an extensive suite of
counterfactual generators. As a major next step, we aim to make our
library as compatible as possible with the popular
\href{https://alan-turing-institute.github.io/MLJ.jl/dev/}{\texttt{MLJ.jl}}
package for machine learning in Julia. We invite the Julia community to
contribute to these goals through usage, open challenge and active
development.

\hypertarget{sec-ack}{%
\section{Acknowledgements}\label{sec-ack}}

We are immensely grateful to the group of TU Delft students who
contributed huge improvements to this package as part of a university
project in 2023: Rauno Arike, Simon Kasdorp, Lauri Kesküll, Mariusz
Kicior, Vincent Pikand. We also want to thank the broader Julia
community for being welcoming and open and for supporting research
contributions like this one. Some of the members of TU Delft were
partially funded by ICAI AI for Fintech Research, an ING---TU Delft
collaboration.

\hypertarget{references}{%
\section*{References}\label{references}}
\addcontentsline{toc}{section}{References}

\hypertarget{refs}{}
\begin{CSLReferences}{1}{0}
\leavevmode\vadjust pre{\hypertarget{ref-altmeyer2023endogenous}{}}%
Altmeyer, Patrick, Giovan Angela, Aleksander Buszydlik, Karol Dobiczek,
Arie van Deursen, and Cynthia Liem. 2023. {``Endogenous {Macrodynamics}
in {Algorithmic} {Recourse}.''} In \emph{First {IEEE} {Conference} on
{Secure} and {Trustworthy} {Machine} {Learning}}.
\url{https://doi.org/10.1109/satml54575.2023.00036}.

\leavevmode\vadjust pre{\hypertarget{ref-antoran2020getting}{}}%
Antorán, Javier, Umang Bhatt, Tameem Adel, Adrian Weller, and José
Miguel Hernández-Lobato. 2020. {``Getting a Clue: {A} Method for
Explaining Uncertainty Estimates.''}
\url{https://arxiv.org/abs/2006.06848}.

\leavevmode\vadjust pre{\hypertarget{ref-arrieta2020explainable}{}}%
Arrieta, Alejandro Barredo, Natalia Diaz-Rodriguez, Javier Del Ser,
Adrien Bennetot, Siham Tabik, Alberto Barbado, Salvador Garcia, et al.
2020. {``Explainable {Artificial Intelligence} ({XAI}): {Concepts},
Taxonomies, Opportunities and Challenges Toward Responsible {AI}.''}
\emph{Information Fusion} 58: 82--115.
\url{https://doi.org/10.1016/j.inffus.2019.12.012}.

\leavevmode\vadjust pre{\hypertarget{ref-becker1996adult}{}}%
Barry Becker, Ronny Kohavi. 1996. {``Adult.''} UCI Machine Learning
Repository. \url{https://doi.org/10.24432/C5XW20}.

\leavevmode\vadjust pre{\hypertarget{ref-blaom2020mlj}{}}%
Blaom, Anthony D., Franz Kiraly, Thibaut Lienart, Yiannis Simillides,
Diego Arenas, and Sebastian J. Vollmer. 2020. {``{MLJ}: {A Julia}
Package for Composable Machine Learning.''} \emph{Journal of Open Source
Software} 5 (55): 2704. \url{https://doi.org/10.21105/joss.02704}.

\leavevmode\vadjust pre{\hypertarget{ref-borch2022machine}{}}%
Borch, Christian. 2022. {``Machine Learning, Knowledge Risk, and
Principal-Agent Problems in Automated Trading.''} \emph{Technology in
Society}, 101852. \url{https://doi.org/10.1016/j.techsoc.2021.101852}.

\leavevmode\vadjust pre{\hypertarget{ref-buolamwini2018gender}{}}%
Buolamwini, Joy, and Timnit Gebru. 2018. {``Gender Shades:
{Intersectional} Accuracy Disparities in Commercial Gender
Classification.''} In \emph{Conference on Fairness, Accountability and
Transparency}, 77--91. {PMLR}.

\leavevmode\vadjust pre{\hypertarget{ref-dandl2023counterfactuals}{}}%
Dandl, Susanne, Andreas Hofheinz, Martin Binder, Bernd Bischl, and
Giuseppe Casalicchio. 2023. {``Counterfactuals: {An} {R} {Package} for
{Counterfactual} {Explanation} {Methods}.''} arXiv.
\url{http://arxiv.org/abs/2304.06569}.

\leavevmode\vadjust pre{\hypertarget{ref-delaney2021uncertainty}{}}%
Delaney, Eoin, Derek Greene, and Mark T. Keane. 2021. {``Uncertainty
{Estimation} and {Out}-of-{Distribution} {Detection} for
{Counterfactual} {Explanations}: {Pitfalls} and {Solutions}.''} arXiv.
\url{http://arxiv.org/abs/2107.09734}.

\leavevmode\vadjust pre{\hypertarget{ref-fan2020interpretability}{}}%
Fan, Fenglei, Jinjun Xiong, and Ge Wang. 2020. {``On Interpretability of
Artificial Neural Networks.''} \url{https://arxiv.org/abs/2001.02522}.

\leavevmode\vadjust pre{\hypertarget{ref-goodfellow2014explaining}{}}%
Goodfellow, Ian J, Jonathon Shlens, and Christian Szegedy. 2014.
{``Explaining and Harnessing Adversarial Examples.''}
\url{https://arxiv.org/abs/1412.6572}.

\leavevmode\vadjust pre{\hypertarget{ref-hoffman1994german}{}}%
Hoffman, Hans. 1994. {``German {Credit Data}.''}
\url{https://archive.ics.uci.edu/ml/datasets/statlog+(german+credit+data)}.

\leavevmode\vadjust pre{\hypertarget{ref-innes2018flux}{}}%
Innes, Mike. 2018. {``Flux: {Elegant} Machine Learning with {Julia}.''}
\emph{Journal of Open Source Software} 3 (25): 602.
\url{https://doi.org/10.21105/joss.00602}.

\leavevmode\vadjust pre{\hypertarget{ref-joshi2019realistic}{}}%
Joshi, Shalmali, Oluwasanmi Koyejo, Warut Vijitbenjaronk, Been Kim, and
Joydeep Ghosh. 2019. {``Towards Realistic Individual Recourse and
Actionable Explanations in Black-Box Decision Making Systems.''}
\url{https://arxiv.org/abs/1907.09615}.

\leavevmode\vadjust pre{\hypertarget{ref-kaggle2011give}{}}%
Kaggle. 2011. {``Give Me Some Credit, {Improve} on the State of the Art
in Credit Scoring by Predicting the Probability That Somebody Will
Experience Financial Distress in the Next Two Years.''} {Kaggle}.
\url{https://www.kaggle.com/c/GiveMeSomeCredit}.

\leavevmode\vadjust pre{\hypertarget{ref-karimi2020survey}{}}%
Karimi, Amir-Hossein, Gilles Barthe, Bernhard Schölkopf, and Isabel
Valera. 2020. {``A Survey of Algorithmic Recourse: Definitions,
Formulations, Solutions, and Prospects.''}
\url{https://arxiv.org/abs/2010.04050}.

\leavevmode\vadjust pre{\hypertarget{ref-karimi2021algorithmic}{}}%
Karimi, Amir-Hossein, Bernhard Schölkopf, and Isabel Valera. 2021.
{``Algorithmic Recourse: From Counterfactual Explanations to
Interventions.''} In \emph{Proceedings of the 2021 {ACM Conference} on
{Fairness}, {Accountability}, and {Transparency}}, 353--62.

\leavevmode\vadjust pre{\hypertarget{ref-karimi2020algorithmic}{}}%
Karimi, Amir-Hossein, Julius Von Kügelgen, Bernhard Schölkopf, and
Isabel Valera. 2020. {``Algorithmic Recourse Under Imperfect Causal
Knowledge: A Probabilistic Approach.''}
\url{https://arxiv.org/abs/2006.06831}.

\leavevmode\vadjust pre{\hypertarget{ref-kaur2020interpreting}{}}%
Kaur, Harmanpreet, Harsha Nori, Samuel Jenkins, Rich Caruana, Hanna
Wallach, and Jennifer Wortman Vaughan. 2020. {``Interpreting
Interpretability: Understanding Data Scientists' Use of Interpretability
Tools for Machine Learning.''} In \emph{Proceedings of the 2020 {CHI}
Conference on Human Factors in Computing Systems}, 1--14.
\url{https://doi.org/10.1145/3313831.3376219}.

\leavevmode\vadjust pre{\hypertarget{ref-krizhevsky2009learning}{}}%
Krizhevsky, A. 2009. {``Learning {Multiple} {Layers} of {Features} from
{Tiny} {Images}.''} In.
\url{https://www.semanticscholar.org/paper/Learning-Multiple-Layers-of-Features-from-Tiny-Krizhevsky/5d90f06bb70a0a3dced62413346235c02b1aa086}.

\leavevmode\vadjust pre{\hypertarget{ref-lakshminarayanan2016simple}{}}%
Lakshminarayanan, Balaji, Alexander Pritzel, and Charles Blundell. 2016.
{``Simple and Scalable Predictive Uncertainty Estimation Using Deep
Ensembles.''} \url{https://arxiv.org/abs/1612.01474}.

\leavevmode\vadjust pre{\hypertarget{ref-laugel2017inversea}{}}%
Laugel, Thibault, Marie-Jeanne Lesot, Christophe Marsala, Xavier Renard,
and Marcin Detyniecki. 2017. {``Inverse {Classification} for
{Comparison}-Based {Interpretability} in {Machine} {Learning}.''} arXiv.
\url{https://doi.org/10.48550/arXiv.1712.08443}.

\leavevmode\vadjust pre{\hypertarget{ref-lecun1998mnist}{}}%
LeCun, Yann. 1998. {``The {MNIST} Database of Handwritten Digits.''}

\leavevmode\vadjust pre{\hypertarget{ref-miller2019explanation}{}}%
Miller, Tim. 2019. {``Explanation in Artificial Intelligence: {Insights}
from the Social Sciences.''} \emph{Artificial Intelligence} 267: 1--38.
\url{https://doi.org/10.1016/j.artint.2018.07.007}.

\leavevmode\vadjust pre{\hypertarget{ref-molnar2020interpretable}{}}%
Molnar, Christoph. 2020. \emph{Interpretable Machine Learning}. {Lulu.
com}.

\leavevmode\vadjust pre{\hypertarget{ref-mothilal2020explaining}{}}%
Mothilal, Ramaravind K, Amit Sharma, and Chenhao Tan. 2020.
{``Explaining Machine Learning Classifiers Through Diverse
Counterfactual Explanations.''} In \emph{Proceedings of the 2020
{Conference} on {Fairness}, {Accountability}, and {Transparency}},
607--17. \url{https://doi.org/10.1145/3351095.3372850}.

\leavevmode\vadjust pre{\hypertarget{ref-oneil2016weapons}{}}%
O'Neil, Cathy. 2016. \emph{Weapons of Math Destruction: {How} Big Data
Increases Inequality and Threatens Democracy}. {Crown}.

\leavevmode\vadjust pre{\hypertarget{ref-pace1997sparse}{}}%
Pace, R Kelley, and Ronald Barry. 1997. {``Sparse Spatial
Autoregressions.''} \emph{Statistics \& Probability Letters} 33 (3):
291--97. \url{https://doi.org/10.1016/s0167-7152(96)00140-x}.

\leavevmode\vadjust pre{\hypertarget{ref-pawelczyk2021carla}{}}%
Pawelczyk, Martin, Sascha Bielawski, Johannes van den Heuvel, Tobias
Richter, and Gjergji Kasneci. 2021. {``Carla: A Python Library to
Benchmark Algorithmic Recourse and Counterfactual Explanation
Algorithms.''} \url{https://arxiv.org/abs/2108.00783}.

\leavevmode\vadjust pre{\hypertarget{ref-pawelczyk2022probabilistically}{}}%
Pawelczyk, Martin, Teresa Datta, Johannes van-den-Heuvel, Gjergji
Kasneci, and Himabindu Lakkaraju. 2022. {``Probabilistically {Robust}
{Recourse}: {Navigating} the {Trade}-Offs Between {Costs} and
{Robustness} in {Algorithmic} {Recourse}.''} \emph{arXiv Preprint
arXiv:2203.06768}.

\leavevmode\vadjust pre{\hypertarget{ref-poyiadzi2020face}{}}%
Poyiadzi, Rafael, Kacper Sokol, Raul Santos-Rodriguez, Tijl De Bie, and
Peter Flach. 2020. {``{FACE}: {Feasible} and Actionable Counterfactual
Explanations.''} In \emph{Proceedings of the {AAAI}/{ACM Conference} on
{AI}, {Ethics}, and {Society}}, 344--50.

\leavevmode\vadjust pre{\hypertarget{ref-rudin2019stop}{}}%
Rudin, Cynthia. 2019. {``Stop Explaining Black Box Machine Learning
Models for High Stakes Decisions and Use Interpretable Models
Instead.''} \emph{Nature Machine Intelligence} 1 (5): 206--15.
\url{https://doi.org/10.1038/s42256-019-0048-x}.

\leavevmode\vadjust pre{\hypertarget{ref-schut2021generating}{}}%
Schut, Lisa, Oscar Key, Rory Mc Grath, Luca Costabello, Bogdan
Sacaleanu, Yarin Gal, et al. 2021. {``Generating {Interpretable
Counterfactual Explanations By Implicit Minimisation} of {Epistemic} and
{Aleatoric Uncertainties}.''} In \emph{International {Conference} on
{Artificial Intelligence} and {Statistics}}, 1756--64. {PMLR}.

\leavevmode\vadjust pre{\hypertarget{ref-slack2021counterfactual}{}}%
Slack, Dylan, Anna Hilgard, Himabindu Lakkaraju, and Sameer Singh. 2021.
{``Counterfactual Explanations Can Be Manipulated.''} \emph{Advances in
Neural Information Processing Systems} 34.

\leavevmode\vadjust pre{\hypertarget{ref-slack2020fooling}{}}%
Slack, Dylan, Sophie Hilgard, Emily Jia, Sameer Singh, and Himabindu
Lakkaraju. 2020. {``Fooling Lime and Shap: {Adversarial} Attacks on Post
Hoc Explanation Methods.''} In \emph{Proceedings of the {AAAI}/{ACM
Conference} on {AI}, {Ethics}, and {Society}}, 180--86.

\leavevmode\vadjust pre{\hypertarget{ref-sturm2014simple}{}}%
Sturm, Bob L. 2014. {``A Simple Method to Determine If a Music
Information Retrieval System Is a {`Horse'}.''} \emph{IEEE Transactions
on Multimedia} 16 (6): 1636--44.
\url{https://doi.org/10.1109/tmm.2014.2330697}.

\leavevmode\vadjust pre{\hypertarget{ref-tolomei2017interpretable}{}}%
Tolomei, Gabriele, Fabrizio Silvestri, Andrew Haines, and Mounia Lalmas.
2017. {``Interpretable {Predictions} of {Tree}-Based {Ensembles} via
{Actionable} {Feature} {Tweaking}.''} In \emph{Proceedings of the 23rd
{ACM} {SIGKDD} {International} {Conference} on {Knowledge} {Discovery}
and {Data} {Mining}}, 465--74.
\url{https://doi.org/10.1145/3097983.3098039}.

\leavevmode\vadjust pre{\hypertarget{ref-upadhyay2021robust}{}}%
Upadhyay, Sohini, Shalmali Joshi, and Himabindu Lakkaraju. 2021.
{``Towards {Robust} and {Reliable Algorithmic Recourse}.''}
\url{https://arxiv.org/abs/2102.13620}.

\leavevmode\vadjust pre{\hypertarget{ref-ustun2019actionable}{}}%
Ustun, Berk, Alexander Spangher, and Yang Liu. 2019. {``Actionable
Recourse in Linear Classification.''} In \emph{Proceedings of the
{Conference} on {Fairness}, {Accountability}, and {Transparency}},
10--19. \url{https://doi.org/10.1145/3287560.3287566}.

\leavevmode\vadjust pre{\hypertarget{ref-varshney2022trustworthy}{}}%
Varshney, Kush R. 2022. \emph{Trustworthy {Machine Learning}}.
{Chappaqua, NY, USA}: {Independently Published}.

\leavevmode\vadjust pre{\hypertarget{ref-verma2020counterfactual}{}}%
Verma, Sahil, John Dickerson, and Keegan Hines. 2020. {``Counterfactual
Explanations for Machine Learning: {A} Review.''}
\url{https://arxiv.org/abs/2010.10596}.

\leavevmode\vadjust pre{\hypertarget{ref-wachter2017counterfactual}{}}%
Wachter, Sandra, Brent Mittelstadt, and Chris Russell. 2017.
{``Counterfactual Explanations Without Opening the Black Box:
{Automated} Decisions and the {GDPR}.''} \emph{Harv. JL \& Tech.} 31:
841. \url{https://doi.org/10.2139/ssrn.3063289}.

\leavevmode\vadjust pre{\hypertarget{ref-xiao2017fashion}{}}%
Xiao, Han, Kashif Rasul, and Roland Vollgraf. 2017. {``Fashion-{MNIST}:
A {Novel} {Image} {Dataset} for {Benchmarking} {Machine} {Learning}
{Algorithms}.''} arXiv. \url{https://doi.org/10.48550/arXiv.1708.07747}.

\leavevmode\vadjust pre{\hypertarget{ref-yeh2009comparisons}{}}%
Yeh, I-Cheng, and Che-hui Lien. 2009. {``The Comparisons of Data Mining
Techniques for the Predictive Accuracy of Probability of Default of
Credit Card Clients.''} \emph{Expert Systems with Applications} 36 (2):
2473--80. \url{https://doi.org/10.1016/j.eswa.2007.12.020}.

\end{CSLReferences}

\end{document}